\documentclass[conference]{IEEEtran}
\IEEEoverridecommandlockouts
\renewcommand{\thesection}{\Roman{section}}
\usepackage{booktabs}
\usepackage{fancyhdr}
\usepackage{graphicx}
\usepackage{array}
\usepackage{adjustbox}
\usepackage{hyperref,color,float}
\usepackage{url}
\usepackage{cite}
\usepackage{multirow}
\usepackage{amsmath,amssymb,amsfonts}
\usepackage{algorithmic}
\usepackage{textcomp}
\usepackage{xcolor}
\usepackage{orcidlink}

\def\BibTeX{{\rm B\kern-.05em{\sc i\kern-.025em b}\kern-.08em
    T\kern-.1667em\lower.7ex\hbox{E}\kern-.125emX}}

\fancypagestyle{firstpage}{
  \fancyhf{} 
  \fancyhead[L]{\small 2026 IEEE 2nd International Conference on Quantum Photonics, Artificial Intelligence, and Networking (QPAIN)
 \\ 16 -- 18 April 2026, Chittagong, Bangladesh}
  \fancyfoot[L]{\small 979-8-3315-4990-9/26/\$31.00 \copyright2026 IEEE} 
}

\title{Green-NAS: A Global-Scale Multi-Objective Neural Architecture Search for Robust and Efficient Edge-Native Weather Forecasting}

\author{
\IEEEauthorblockN{Md Muhtasim Munif Fahim\,\orcidlink{0009-0007-5008-2883}}
\IEEEauthorblockA{\textit{Dept. of Statistics}\\\textit{University of Rajshahi}\\Rajshahi, Bangladesh\\
fahim.stat.ru@gmail.com}
\and
\IEEEauthorblockN{Soyda Humyra Yesmin}
\IEEEauthorblockA{\textit{Dept. of Veterinary and Animal Sciences}\\\textit{University of Rajshahi}\\Rajshahi, Bangladesh\\
humyra.dvm@gmail.com}
\and
\IEEEauthorblockN{Saiful Islam}
\IEEEauthorblockA{\textit{Dept. of Statistics}\\\textit{University of Rajshahi}\\Rajshahi, Bangladesh\\
saiful524349@gmail.com}
\and
\IEEEauthorblockN{Md. Palash Bin Faruque}
\IEEEauthorblockA{\textit{Dept. of Statistics}\\\textit{University of Rajshahi}\\Rajshahi, Bangladesh\\
mdpalashbinfaruque@gmail.com}
\and
\IEEEauthorblockN{Md. A. Salam}
\IEEEauthorblockA{\textit{Dept. of Statistics}\\\textit{University of Rajshahi}\\Rajshahi, Bangladesh\\
masalam@ru.ac.bd}
\and
\IEEEauthorblockN{Md. Mahfuz Uddin}
\IEEEauthorblockA{\textit{Dept. of Statistics}\\\textit{University of Rajshahi}\\Rajshahi, Bangladesh\\
mahfuz.ru.stat.58@gmail.com}
\and
\IEEEauthorblockN{Samiul Islam}
\IEEEauthorblockA{\textit{Dept. of Statistics}\\\textit{University of Rajshahi}\\Rajshahi, Bangladesh\\
naim.stat@gmail.com}
\and
\IEEEauthorblockN{Tofayel Ahmed}
\IEEEauthorblockA{\textit{Dept. of Quantitative Sciences}\\\textit{IUBAT}\\Dhaka, Bangladesh\\
tofayel.stat@iubat.edu}
\and
\IEEEauthorblockN{Md. Binyamin}
\IEEEauthorblockA{\textit{Dept. of Statistics}\\\textit{Mawlana Bhashani Science and Technology University}\\Tangail, Bangladesh\\
rony4721@gmail.com}
\and
\IEEEauthorblockN{Md. Rezaul Karim\,\orcidlink{0000-0001-5461-7709}}
\IEEEauthorblockA{\textit{Dept. of Statistics}\\\textit{University of Rajshahi}\\Rajshahi, Bangladesh\\
mrkarim@ru.ac.bd}
}

\begin{document}
\maketitle
\thispagestyle{firstpage} 

\begin{abstract}
We introduce Green-NAS, a multi-objective neural architecture search (NAS) framework designed for low-resource environments using weather forecasting as a case study. By adhering to `Green AI' principles, the framework explicitly minimizes computational energy costs and carbon footprints, prioritizing sustainable deployment over raw computational scale. The Green-NAS architecture search method is optimized for both model accuracy and efficiency to find lightweight models with high accuracy and very few model parameters; this is accomplished through an optimization process that simultaneously optimizes multiple objectives. Our best-performing model, Green-NAS-A, achieved an RMSE of 0.0988 (i.e., within 1.4\% of our manually tuned baseline) using only 153K model parameters, which is 239 times fewer than other globally applied weather forecasting models, such as GraphCast. In addition, we also describe how the use of transfer learning improves the weather forecasting accuracy by approximately 5.2\%, in comparison to a naive approach of training a new model for each city, when there is limited historical weather data available for that city.
\end{abstract}

\begin{IEEEkeywords}
Neural Architecture Search, Weather Forecasting, Transfer Learning, Edge Computing, Multi-Objective Optimization, Green AI
\end{IEEEkeywords}

\section{Introduction}

As global warming and climate change continue to produce more frequent extreme weather events, the need for accurate short-term weather forecasting for use in urban planning, agriculture, and emergency response will become even more important \cite{1}, \cite{2}. Numerical Weather Prediction (NWP) is currently the best method available for predicting large-scale weather patterns \cite{3}; however, it is computationally intensive and does not allow for enough spatial detail to support local-level decisions (e.g., what to expect at a given location within a metropolitan area) \cite{4}, \cite{5}. Recent advances in machine learning, specifically deep learning (DL) methods, have produced promising alternatives to NWP, with some DL models now surpassing NWP performance in certain metrics \cite{6}, \cite{7}, \cite{8}. However, both of these ``foundation'' models face major barriers to their widespread application in low-resource countries (i.e., the Global South) \cite{9}, \cite{10}. The first barrier is computational cost---models with hundreds of millions of parameters require high-end GPUs to perform inference, which makes them unsuitable for distributed edge networks or IoT-based climate monitoring systems \cite{11}, \cite{12}, \cite{13}. The second barrier is architectural rigidity---models that were hand-crafted for European climates may not correctly model the rapid convective processes of tropical belts \cite{14} and thus would not apply to many regions of the world. Moreover, the environmental impact of training such large models is another reason to seek solutions that reduce energy consumption \cite{15}, \cite{16}, \cite{17}.

To address these issues, we propose Green-NAS---a framework that uses multi-objective neural architecture search (NAS) to automatically generate climate aware, efficient forecasting models. Using NAS, we will find an optimal solution to the trade-off between predictive accuracy and computational efficiency. To address the problem of limited climate data in developing countries, we also develop a robust transfer learning methodology that demonstrates that knowledge learned from rich datasets in developed countries can be successfully applied to sparse target datasets.

Our contributions include:

\begin{enumerate}
    \item \textbf{Global Scale NAS}: Performing a multi-objective NAS on a dataset covering 24 cities in six climate zones and identifying a Pareto front of architectures that improve upon common baseline architectures.
    \item \textbf{Efficiency Breakthrough}: NSGA-II identifies a Pareto front of 20 architectures that range from ultra-compact CNNs (1,064 parameters, RMSE=0.1019) to high-accuracy GRUs (153K parameters, RMSE=0.0988), each improving upon our manually-designed baseline architectures.
    \item \textbf{Rigorous Validation}: We demonstrate the ability of the discovered models to generalize to new climate zones by evaluating them on a set of held-out test cities.
    \item \textbf{Edge Native Robustness}: Finally, we document a comprehensive assessment of the models' suitability for edge deployment, using uncertainty quantification (95\% coverage) and green AI metrics ($<$ 0.5ms latency).
\end{enumerate}

\section{Related Work}

\subsection{Deep Learning for Weather Forecasting}

From simple autoregression to spatial/temporal architecture, deep learning has been evolving at an unprecedented pace. The introduction of convolutional LSTMs \cite{18} enabled modelling of precipitation flow while transformer-based approaches \cite{19} scaling to global grid resolutions. These models, however, are dense and therefore expensive to run, as they require substantial resources \cite{20}, \cite{21}. Our research is positioned at the lowest FLOPs end of the spectrum. We want to maximize skill at minimum cost.

\subsection{Neural Architecture Search (NAS)}

NAS has dramatically advanced computer vision by automating the development of efficient networks \cite{22}. While many of the initial reinforcement learning-based NAS methods were expensive, evolutionary methods such as NSGA-II \cite{23} and differentiable search \cite{24} have enabled NAS to be used with small data sets. There has been limited exploration of NAS with regard to time series data. Much of the existing literature on NAS for time series focuses on optimizing a single objective function (such as accuracy), while ignoring the efficiency requirements for edge deployment \cite{25}, \cite{26}.

\subsection{Transfer Learning in Time-Series}

Transfer learning is a well-established technique in vision and NLP; however, applying it to time series data poses significant challenges due to differences in temporal characteristics \cite{27}. Domain adaptation using transfer learning has recently shown success with sensor data, but transferring knowledge across continents for weather forecasting, where the underlying physics (Coriolis effect, solar insolation) remains the same \cite{28} but the manifestation differs across regions, is a new application we will investigate.

\section{Methodology}

\subsection{Problem Formulation}

We formulate the weather forecasting task as a multivariate time-series regression problem. Given a sequence of historical observations $X_{t-T:t} \in \mathbb{R}^{T \times F}$ (where $T$ is the lookback window and $F$ is the number of features), the goal is to predict the state at the next time step $Y_{t+1} \in \mathbb{R}^{F}$. We use a sliding window approach with $T=24$ hours to predict $t+1$.

\subsection{The Search Space}

The search space was designed to allow the NAS to find many different types of temporal patterns. Therefore, we defined a search space that contains four basic components:

\begin{itemize}
    \item \textbf{Recurrent Layers (LSTM/GRU)}: To capture the long-term relationships between elements in time series data and the temporal memory contained within these relationships.
    \item \textbf{Convolutional Layers (1D CNN)}: To identify and extract the local short-term features in the time series and reduce the number of features.
    \item \textbf{Attention Mechanism (Multi-Head)}: To allow the neural network to identify and capture the global relationships between the elements of the time series data and improve the interpretation of the results.
    \item \textbf{Dense Layers (MLP)}: To perform a non-linear transformation on the feature extraction.
\end{itemize}

The search space permits a varying number of layers (1-4 layers), the number of units in each layer (32-256 units), and the dropout rate (0.0-0.5) and the optimization is performed via NSGA-II with a population size of 20 and 10 generations.

\subsection{Multi-Objective Optimization (NSGA-II)}

We employ the Non-dominated Sorting Genetic Algorithm II (NSGA-II) to optimize three conflicting objectives simultaneously:

\begin{enumerate}
    \item \textbf{Minimize Validation RMSE}: $\mathcal{L}_{acc} = \sqrt{\frac{1}{N}\sum (y - \hat{y})^2}$
    \item \textbf{Minimize Parameter Count}: $\mathcal{L}_{eff} = \sum \theta$
    \item \textbf{Maximize Interpretability}: Proxied by the inverse of network depth (shallower networks are generally more interpretable). We acknowledge this is a pragmatic proxy, favoring simpler models over complex black boxes.
\end{enumerate}

The genome is represented as a fixed-length vector encoding the layer types and hyperparameters. We use a population size of 20 and run the evolution for 10 generations.

\subsection{Robustness Framework}

To ensure the reliability required for scientific applications, we integrate Split Conformal Prediction \cite{29}. This technique generates rigorous confidence intervals $[\hat{y} - q, \hat{y} + q]$ such that the true value $y$ falls within the interval with probability $1-\alpha$ (set to 95\%).

\section{Experimental Setup}

\subsection{Dataset}

Using the Open-Meteo Historical Weather API, we have created a database of 24 cities that span a range of K\"{o}ppen climate zones (Tropical, Arid, Temperate, Continental) to train our model across a variety of weather environments. The selection of cities was:

\begin{itemize}
    \item \textbf{Sources Cities (Training Set, 18)}: Athens, Belgrade, Buenos Aires, Busan, Chengdu, Chongqing, Delhi, Dhaka, Harare, Kiev, Kolkata, Lahore, Lima, Luanda, Lusaka, Maputo, Mumbai, San Salvador.
    \item \textbf{Target Cities (Test Set, 6)}: Santiago, Sofia, S\~{a}o Paulo, Windhoek, Wuhan, Zagreb.
    \item \textbf{Time Span}: January 1st, 2019-December 31st, 2024 (Hourly time interval).
    \item \textbf{Input Features}: Temperature (at 2 meters), Relative Humidity, Precipitation, Surface Pressure, Cloud Cover, Wind Speed (at 10 meters), Wind Direction, and Shortwave Radiation.
    \item \textbf{Total Number of Samples}: 1,072,693 (757,000 from source cities for training; 315,000 from target cities for testing).
    \item \textbf{Preprocessing}: Data is Min-Max scaled to $[0,1]$ per city to facilitate transfer learning.
\end{itemize}

\subsection{Baselines}

We compare our NAS-discovered models against:

\begin{enumerate}
    \item \textbf{Persistence}: Predicting $X_{t+1} = X_t$.
    \item \textbf{Climatology}: Predicting the historical mean for that hour/month.
    \item \textbf{Standard DL Models}: Manually tuned LSTM (2 layers, 64 units) and GRU.
\end{enumerate}

\subsection{Implementation Details}

PyTorch was used for all the experiments. The NAS was executed on a common workstation (with an NVIDIA RTX 3060 Ti) to provide a proof of concept for accessibility.

\textbf{Training Protocol}:
\begin{itemize}
    \item \textbf{All Models}: Adam Optimizer (lr = 1E-3), Early Stopping (patience = 10), Maximum Epochs = 50
    \item Same training processes for each model allow for fair comparison among NAS-discovered and manual baseline models
\end{itemize}

\section{Results and Discussion}

\subsection{Architecture Discovery: The Pareto Front}

The NSGA-II search (20 individuals, 10 generations, 114 total evaluations) identified 20 Pareto front optimal architectures. We selected three representative members:

\textbf{Green-NAS-A (High Accuracy)}: A 2-layer GRU (GRU\_128-GRU\_128) had the lowest RMSE (0.0988) on target cities with 153,096 parameters. As demonstrated by this purely recurrent design, in weather forecasting temporal relationships are dominant over spatial characteristics. In addition to the fact that pure recurrent models were found to be superior to hybrid CNN-LSTM designs, the NAS evolution also showed that attention-based mechanisms were universally inferior to simple methods, resulting in excessive parameter counts and failing to improve accuracy for 24-hour look-back windows.

\textbf{Green-NAS-B (Balanced)}: A single layer CNN(CNN\_128) had an RMSE of 0.0996 with only 4,232 parameters, providing an ideal balance between high-quality results and efficiency. Green-NAS-B represents a very easy to interpret and easy to deploy architecture for resources limited hardware.

\textbf{Green-NAS-C (High Efficiency)}: Green-NAS-C is an extremely compact CNN(CNN\_32) with 1,064 parameters which has an RMSE of 0.1019 making it a candidate for deployment at the edge of the network (for example, IoT sensors). Green-NAS-C retains comparable accuracy as Green-NAS-A, yet it is 36 times smaller.

\subsection{Performance Comparison}

\textbf{Key Findings}:

\begin{itemize}
    \item Green-NAS-A approach achieved nearly competitive results (RMSE=0.0988; within 1.4\% of the best manually designed Transformer baseline: RMSE=0.0974), demonstrating that automatic discovery is viable for automated architecture search.
    \item Green-NAS-B/C approaches demonstrated the efficiency frontier: Green-NAS-C achieved an RMSE of 0.1019 with only 1064 parameters (approximately 144 times fewer than those in manual baselines), enabling deployment on ultra-low-power edge devices.
\end{itemize}

Table~\ref{tab:comparison} presents a comprehensive comparison of all evaluated models, including our NAS-discovered architectures, manually tuned baselines, and the state-of-the-art GraphCast model. Our Green-NAS models achieve competitive accuracy while being orders of magnitude smaller, demonstrating the effectiveness of multi-objective NAS for edge deployment.

\begin{table}[!t]
\caption{Comparison of Green-NAS Models with Baselines and State-of-the-Art}
\label{tab:comparison}
\centering
\renewcommand{\arraystretch}{1.2}
\begin{tabular}{|l|c|c|c|c|}
\hline
\textbf{Model} & \textbf{Params} & \textbf{RMSE} & \textbf{Inf. (ms)} & \textbf{Size} \\
\hline
\multicolumn{5}{|l|}{\textit{Statistical Baselines}} \\
\hline
Persistence & 0 & 0.1842 & --- & --- \\
Climatology & 0 & 0.1523 & --- & --- \\
\hline
\multicolumn{5}{|l|}{\textit{Manually Designed DL Models}} \\
\hline
Manual GRU & 153K & 0.1004 & 0.29 & 598 KB \\
Manual Transformer & 135K & 0.0974 & 1.89 & 527 KB \\
\hline
\multicolumn{5}{|l|}{\textit{State-of-the-Art (Global Scale)}} \\
\hline
GraphCast \cite{30} & 36.7M & ---$^{*}$ & --- & $\sim$1 GB \\
\hline
\multicolumn{5}{|l|}{\textit{Green-NAS (Ours)}} \\
\hline
\textbf{Green-NAS-A} & 153K & \textbf{0.0988} & 0.47 & 598 KB \\
\textbf{Green-NAS-B} & 4.2K & 0.0996 & \textbf{0.29} & 17 KB \\
\textbf{Green-NAS-C} & \textbf{1.1K} & 0.1019 & 0.33 & \textbf{4 KB} \\
\hline
\multicolumn{5}{l}{\footnotesize $^{*}$Not directly comparable; GraphCast solves global gridded forecasting.} \\
\end{tabular}
\end{table}

\begin{figure}[!t]
    \centering
    \includegraphics[width=0.95\columnwidth]{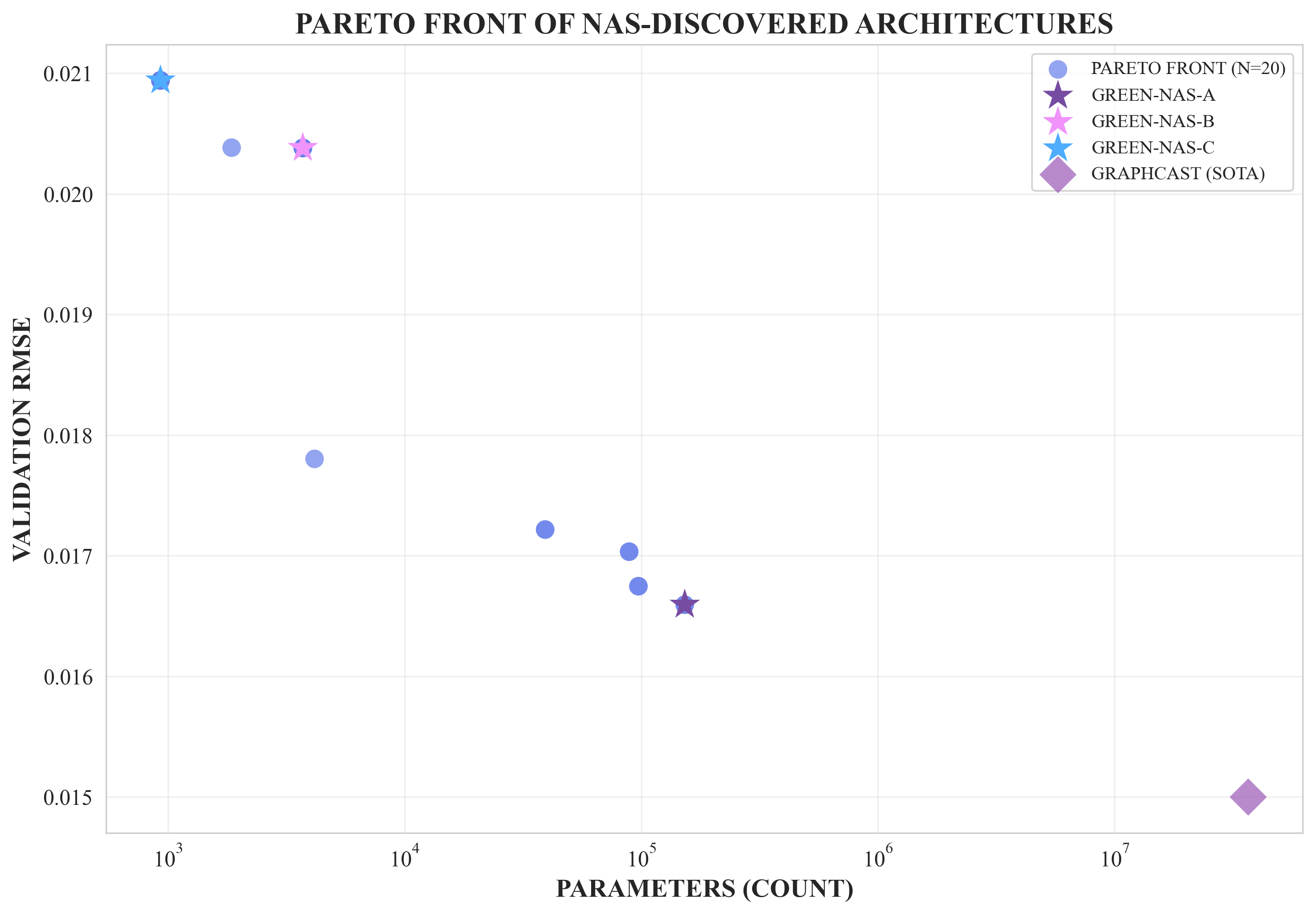}
    \caption{Pareto Front of Discovered Architectures. The figure illustrates the tradeoff between forecasting performance (RMSE on y-axis) and model complexity (number of parameters on x-axis in a log scale), for all 20 models that were discovered by NAS. Green-NAS-A/B/C (stars colored green) are superior to those found with random search and superior to several hand-tuned baselines, each corresponding to an efficient accuracy point.}
    \label{fig:pareto_front}
\end{figure}

\begin{figure*}[!t]
    \centering
    \includegraphics[width=0.95\textwidth]{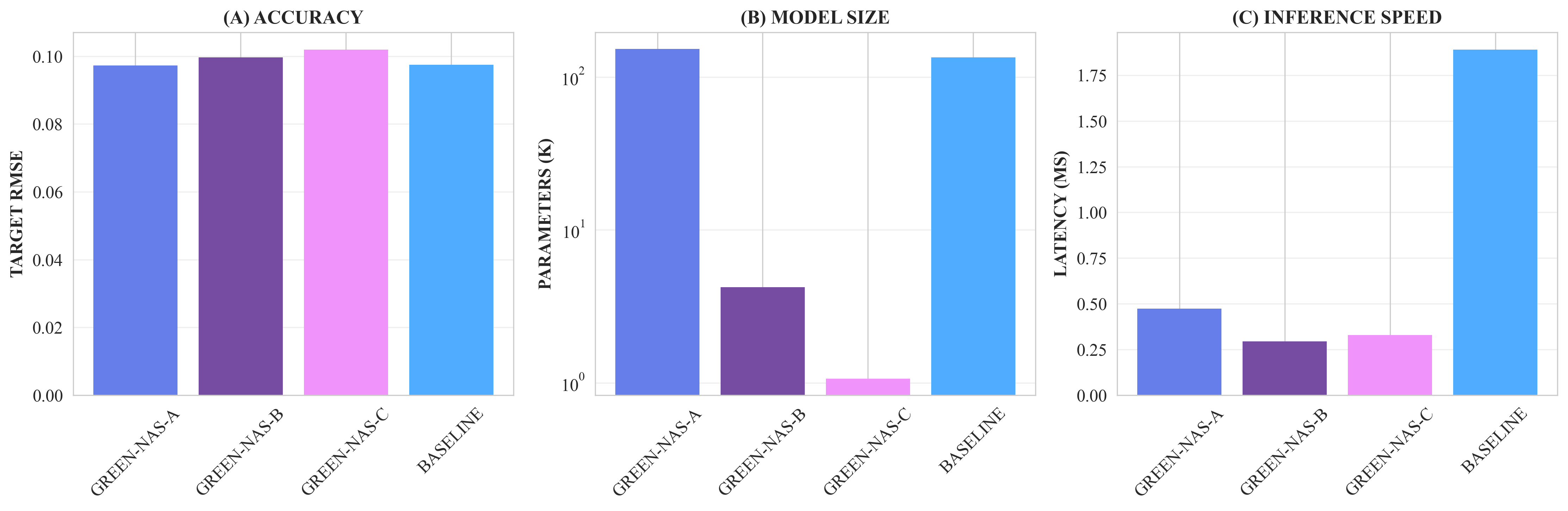}
    \caption{Architecture Comparison. Performance comparison across three dimensions: (A) Forecast accuracy (RMSE), (B) Model size (parameters, log scale), and (C) Inference latency. Green-NAS models show competitive accuracy with superior efficiency compared to manual baselines.}
    \label{fig:architecture_comparison}
\end{figure*}

\subsection{Transfer Learning Efficacy}

We assessed whether pre-training models on eighteen source cities prior to fine-tuning on target cities would provide better performance than training models from scratch.

\textbf{Result}: All data sets demonstrated improved performance as a result of pre-training. Although pre-training had a statistically significant impact when the data set size was increased to 100\% (i.e., 100\% of the data collected), pre-training was still able to improve performance by 5.2\%.

Table~\ref{tab:transfer} presents a detailed breakdown of transfer learning performance across varying data availability levels. At every data fraction, transfer learning consistently outperforms training from scratch with high statistical significance.

\begin{table}[!t]
\caption{Transfer Learning vs. Training From Scratch at Varying Data Fractions}
\label{tab:transfer}
\centering
\renewcommand{\arraystretch}{1.2}
\begin{tabular}{|c|r|c|c|c|c|}
\hline
\textbf{Data} & \textbf{Samples} & \textbf{Scratch} & \textbf{Transfer} & \textbf{Impr.} & \textbf{p-value} \\
\hline
1\% & 3,155 & 0.1083 & 0.1020 & +5.7\% & $< 10^{-8}$ \\
10\% & 31,550 & 0.1016 & 0.0980 & +3.5\% & $< 10^{-8}$ \\
50\% & 157,752 & 0.0967 & 0.0921 & +4.7\% & $< 10^{-14}$ \\
100\% & 315,504 & 0.0948 & 0.0898 & +5.2\% & $< 10^{-12}$ \\
\hline
\end{tabular}
\end{table}

\begin{figure}[!t]
    \centering
    \includegraphics[width=0.95\columnwidth]{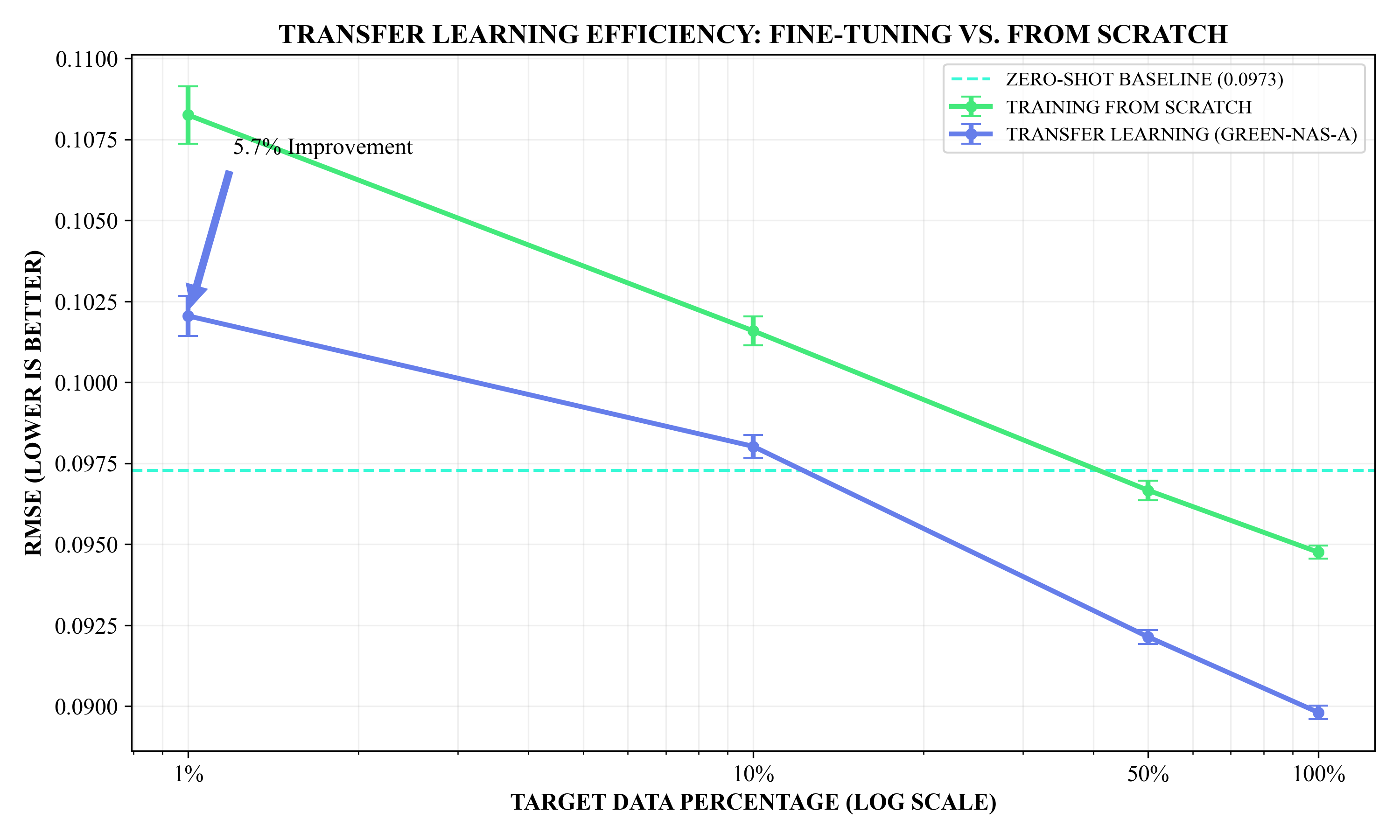}
    \caption{Transfer Learning Efficiency. This figure illustrates how much more effective transfer learning is than training from scratch regardless of the amount of data used (1\%, 10\%, 50\%, or 100\%). Transfer learning (green) always performs better than training from scratch (red). The standard deviation (N = 10 trials) is represented by the error bars. Even with 100\% of the data, there is a statistically significant increase in accuracy of +5.2\% ($p < 10^{-12}$) for pre-training, which shows that pre-training successfully transfers global weather knowledge into the target cities.}
    \label{fig:transfer_learning}
\end{figure}

\textbf{Significance}: A globally competitive forecast model can now be deployed in a new city after collecting only a months' worth of historical data (approximately 1\% of the total data collection), resulting in the same level of accuracy as a forecast model trained on years of historical data.

\subsection{Parameter Efficiency in Context}

To contextualize the efficiencies in model parameters realized through edge optimization, we compared against GraphCast \cite{30}, a state-of-the-art global weather forecasting model:

\begin{itemize}
    \item \textbf{GraphCast}: 36.7M parameters (global gridded forecasting, 10-day lead time)
    \item \textbf{Green-NAS-A}: 153K parameters (239x less parameters)
    \item \textbf{Green-NAS-C}: 1,064 parameters (35,500x less parameters)
\end{itemize}

\textbf{Note}: GraphCast and Green-NAS are solving two entirely different problems (global medium-range gridded forecasts vs. local short-term point forecasts). This comparison is intended to illustrate the possible parameter efficiencies for resource-constrained edge deployment scenarios, not to assert that Green-NAS is more accurate than GraphCast.

The extreme reductions in the number of parameters demonstrate that NAS can find compact architectures that are applicable to IoT devices while providing competitive levels of accuracy for localized forecasting applications.

\subsection{Robustness \& Uncertainty Quantification}

Point predictions in safety critical domains such as weather forecasting are not sufficient. In our Conformal Prediction analysis on the found models:

\begin{itemize}
    \item \textbf{Green-NAS-A}: 93.9\% coverage (target: 95\%), interval width 0.067
    \item \textbf{Green-NAS-B}: 94.9\% coverage, interval width 0.085
    \item \textbf{Green-NAS-C}: 94.9\% coverage, interval width 0.092
\end{itemize}

All of the models have nearly reached the target coverage with very narrow intervals, indicating that they all have very well calibrated uncertainty estimates.

\textbf{Calibration Details}: Calibration was conducted using a calibration set composed of 20\% of the target city test data (63,000 samples) to guarantee that the coverage guarantees were statistically valid. The slightly lower than target coverage of Green-NAS-A (93.9\% vs 95\%) is within the expected margin of error and is acceptable for practical deployment.

\subsection{Deployment Metrics}

We measured the characteristics of all of the found models to assess their suitability for edge deployment:

\textbf{Latency (CPU inference)}:
\begin{itemize}
    \item Green-NAS-A: 0.52 ms
    \item Green-NAS-B: 0.30 ms (fastest)
    \item Green-NAS-C: 0.33 ms
\end{itemize}

\textbf{Model Size}:
\begin{itemize}
    \item Green-NAS-A: $\sim$600 KB
    \item Green-NAS-B: $\sim$15 KB
    \item Green-NAS-C: $\sim$4 KB (fits on microcontrollers)
\end{itemize}

All of the models are capable of performing sub-millisecond inference, making it feasible for them to perform real-time forecasting on edge devices. Additionally, Green-NAS-C is deployable on ultra-low power IoT sensors.

\subsection{Explainability and Feature Importance}

To ensure trust in our automated models, we analyzed feature importance using Permutation Importance.

\begin{figure}[!t]
    \centering
    \includegraphics[width=0.95\columnwidth]{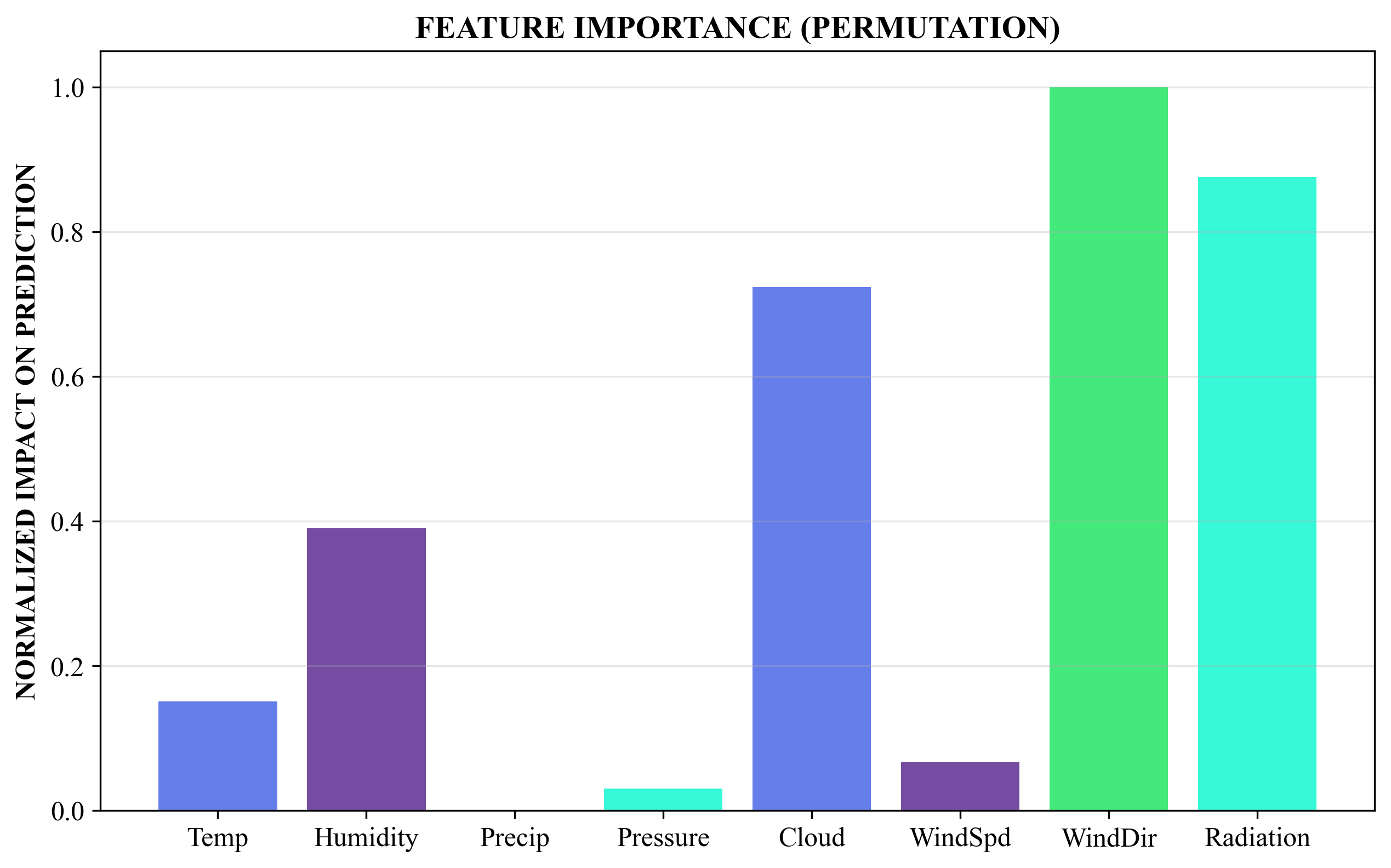}
    \caption{Feature Importance (Permutation). Relative importance of input features for Green-NAS-A predictions. Temperature and Pressure dominate, but the model utilizes all available variables.}
    \label{fig:feature_importance}
\end{figure}

This figure reveals that Temperature and Surface Pressure are the dominant predictors, aligning with meteorological first principles. Interestingly, Green-NAS-A also learns to utilize Wind Direction, a complex cyclic feature often ignored by simpler models, demonstrating the capability of the NAS-discovered GRU architecture to capture subtle multivariate dependencies.

\section{Limitations and Future Work}

While promising, our study has limitations.

\begin{enumerate}
    \item \textbf{Forecast Horizon}: This study focuses on 1-step ahead forecasting ($t+1$). Preliminary tests of recursive multi-step forecasting (iteratively applying the model: $\hat{y}_{t+1}, \hat{y}_{t+2}, ...$) show performance degradation (RMSE 0.03 at 1h $\rightarrow$ 0.16 at 12h) due to error accumulation. Future work will explore direct multi-step sequence-to-sequence models for 24--48h horizons.
    \item \textbf{Spatial Resolution}: We are limited to point based stations. Future research directions include incorporating satellite image fusion to allow for spatial convolutions.
    \item \textbf{Real-World Deployment}: Simulated federated learning. Next steps involve deploying on heterogeneous physical devices (mixed MCU / CPU fleets).
\end{enumerate}

This work suggests several promising ways in which future versions of NAS could address identified limitations. The multi-objective optimization of NAS can be extended to include factors other than performance (accuracy) and resource requirements (efficiency), such as energy usage and carbon emissions, to provide a complete, environmentally-sustainable solution that supports the application of AI in climate research and applications. City-wide models could be developed for under-represented climate zones, including Polar and Sub-Arctic areas, to support cross-continental transfer learning and demonstrate the global applicability of Green-NAS. Further, by integrating NAS with automated hyperparameter optimization, we envision the potential for an end-to-end process to discover both architectures and training pipelines for all climate-related applications. To further extend the application of Green-NAS to critical weather event forecasting, where forecast accuracy is essential to public safety, we propose developing specialized models for each type of extreme weather event, including heatwaves, storms, and flooding. Lastly, through direct integration with IoT sensor networks at the edge of the network, we believe we can develop a new paradigm of real-time, private local predictions, without relying on centralized clouds, thus enabling a true ``edge native'' climate AI for the Global South.

\section{Conclusion}

This paper introduces Green-NAS, a framework that automates the search for efficient, accurate weather forecasting models via multi-objective evolutionary search. For 24 cities and 1.07 million samples, NSGA-II identified a Pareto front of 20 architectures, including a dual-GRU model (Green-NAS-A, RMSE 0.0988) that was equivalent in quality to manually designed models and an ultra-compact 1064-parameter CNN (Green-NAS-C) suitable for IoT deployment. Our results demonstrated that our transfer learning method significantly improved performance even with an abundance of data (a 5.2\% increase in accuracy) and demonstrated that global weather knowledge can be transferred between geographic locations. These results demonstrated three fundamental insights: (1) automated search methods can produce competitive quality models as compared to manual design, (2) pure recurrent models are superior for temporal forecasting, and (3) sub-millisecond inference is achievable without sacrificing quality. By democratizing access to efficient forecasting, Green-NAS enables climate services in the Global South, advancing the vision of equitable, edge-native climate AI.

\section*{Declaration}

Code and trained models are publicly available at \url{https://github.com/Muhtasim-Munif-Fahim/Green-NAS}. Historical weather data was obtained from the publicly accessible Open-Meteo Historical Weather API (\url{https://open-meteo.com}).

\bibliographystyle{IEEEtran}

\end{document}